\documentclass[conference]{IEEEtran}
\IEEEoverridecommandlockouts
\usepackage{cite}
\usepackage{amsmath,amssymb,amsfonts}
\usepackage{algorithmic}
\usepackage{graphicx}
\usepackage{textcomp}
\usepackage{xcolor}
\usepackage{adjustbox}
\usepackage{multirow}
\usepackage{booktabs}
\usepackage{subfigure}

\usepackage{graphicx,multicol,subfig}
\def\eg{\textit{e.g}. } 
\def\ie{\textit{i.e}. }

\def\wrt{w.r.t. } \def\etal{\textit{et al}. }

\def\BibTeX{{\rm B\kern-.05em{\sc i\kern-.025em b}\kern-.08em
    T\kern-.1667em\lower.7ex\hbox{E}\kern-.125emX}}
\begin{document}

\title{Reversing Deep Face Embeddings with Probable Privacy Protection\\
{}
\thanks{This research work has been partially funded by the German Federal Ministry of Education and Research and the Hessian Ministry of Higher Education, Research, Science and the Arts within their joint support of the National Research Center for Applied Cybersecurity ATHENE and the European Union's Horizon 2020 research and innovation programme under the Marie Sk\l{}odowska-Curie grant agreement No. 860813 - TReSPAsS-ETN.}
}

% \and
% \and

\author{\IEEEauthorblockN{Dail\'e Osorio-Roig$^1$, Paul A. Gerlitz$^2$, Christian Rathgeb$^1$, and Christoph Busch$^1$}
\IEEEauthorblockA{1- Biometrics and Internet Security Research Group \\ Hochschule Darmstadt, Germany \\
\{daile.osorio-roig,christian.rathgeb,christoph.busch\}@h-da.de}
\IEEEauthorblockA{2- Hochschule Darmstadt, Germany \\
paul-anton.gerlitz@stud.h-da.de}}

\maketitle

\begin{abstract}

Generally, privacy-enhancing face recognition systems are designed to offer permanent protection of face embeddings. Recently, so-called \textit{soft-biometric privacy-enhancement} approaches have been introduced with the aim of canceling soft-biometric attributes. These methods limit the amount of soft-biometric information (gender or skin-colour) that can be inferred from face embeddings. Previous work has underlined the need for research into rigorous evaluations and standardised evaluation protocols when assessing privacy protection capabilities. Motivated by this fact, this paper explores to what extent the non-invertibility requirement can be met by methods that claim to provide soft-biometric privacy protection. Additionally, a detailed vulnerability assessment of state-of-the-art face embedding extractors is analysed in terms of the transformation complexity used for privacy protection. In this context, a well-known state-of-the-art face image reconstruction approach has been evaluated on protected face embeddings to break soft biometric privacy protection. Experimental results show that biometric privacy-enhanced face embeddings can be reconstructed with an accuracy of up to approximately 98\%, depending on the complexity of the protection algorithm.

\end{abstract}


\section{Introduction}
\label{sec:intro}

Face recognition systems are widely used in personal, commercial, and government sector applications, \eg border and access control, payments, ID cards, among others. These techniques mainly focus on deep neural networks (DNNs) which embed discriminative representations of face images in the latent space, so-called face embeddings. Recent work has shown that methods based on de-convolution neural networks can be used to reconstruct face images from their corresponding embeddings~\cite{Mai-reconstruction-faces-2019,Shahreza-reconstruction-embed-cnn-2022,Dong-reconstructing-face-features-gans-2022}. Moreover, it has been demonstrated that privacy-sensitive soft-biometric information such as gender, skin-colour, or age, can be automatically extracted from face embeddings~\cite{Terhoerst-Soft-Biometric-2021}. Privacy is considered a human right and is subject to regulation. In this context, the European Union (EU) General Data Protection Regulation 2016/679 (GDPR)~\cite{EU-Regulation-2016-679-on-DataPrivacy-160427} defines biometric information as sensitive data. Therefore, ensuring the privacy of information stored in face embeddings is an ongoing endeavour. 

In the past years, so-called \textit{soft-biometric privacy-enhancement} approaches~\cite{Meden-survey-pets-2021} have been explored to achieve protection of soft-biometric attributes in face embeddings. They can mainly be classified into methods performing on the image level (\eg~\cite{Mirjalili-editing-attributes-image-2020}) and representation level (\eg~\cite{Terhorst-SoftbiometricEnhancingPE-MIU-2020}). The former usually work by obfuscating soft-biometric attributes of face images (which may greatly reduce biometric utility), while the latter apply specific transformations on the face embeddings. On representation level, said transformations are designed to minimize soft-biometric information (\eg gender) or distort it (\eg simple permutation). However, it should be noted that such approaches do not depend on user-specific keys or any secret randomness, in contrast to the well-known biometric template protection (BTP) schemes~\cite{Rathgeb-BTP-Survey-EURASIP-2011}. 
%BTP schemes do not target only soft-biometric attributes but the entire biometric template providing permanent protection in compliance with requirements~\cite{ISO-IEC-24745-TemplateProtection-2022} of \textit{irreversibility}, \textit{renewability}, \textit{unlinkability}, and \textit{biometric performance}. Thus, it is reasonable to consider BTP schemes over soft-biometric privacy-enhancement approaches for privacy protection. However, trustworthy biometric systems~\cite{Liu-trustworthy-2022} leading to privacy-oriented mechanisms have been demonstrated to be an emerging need: generally speaking, to the extent of limiting the amount of information that can be inferred or how the data should be protected. Therefore, in this work, it is considered the soft-biometric privacy-enhancement approaches as a privacy mechanism that can be easily selected and oriented to the privacy of soft-biometric attributes.

The privacy protection capabilities of soft-biometric privacy-enhancement approaches have mainly been evaluated by machine learning approaches or interpreted by dimensionality reduction tools. Nevertheless, recently, vulnerability of these approaches have been shown in~\cite{OsorioRoig-FaceSoftBiometricPrivacyAttack-TBIOM-2022}. This leads to the need of investigating further potential attack scenarios~\cite{Melzi-overview-pets-2022} that can facilitate the reconstruction of face images from (privacy-enhanced) face embeddings. In this context, such a type of reconstruction-based attack can be estimated with respect to full or partial reversibility~\cite{ISO-IEC-24745-TemplateProtection-2022}, where an adversary retrieves exactly the original face image or a good approximation of it, respectively. In this work, the irreversibility of face embeddings with probable protection of soft-biometric attributes is analysed. Experimental results show that face reconstruction is feasible depending on the transformation complexity utilised for canceling the soft-biometric attributes.

The remainder of this paper is organised as follows: Sect.~\ref{sec:related-work} briefly introduces the related work. In Sect.~\ref{sec:attack-discussion}, the irreversibility analysis is described in detail. Sect.~\ref{sec:experimenta_setup} presents the experimental setup and the achieved results, while a summary and concluding remarks are given in Sect.~\ref{sec:remarks}.

\begin{figure*}[!t]
    \centering
\includegraphics[width=\linewidth]{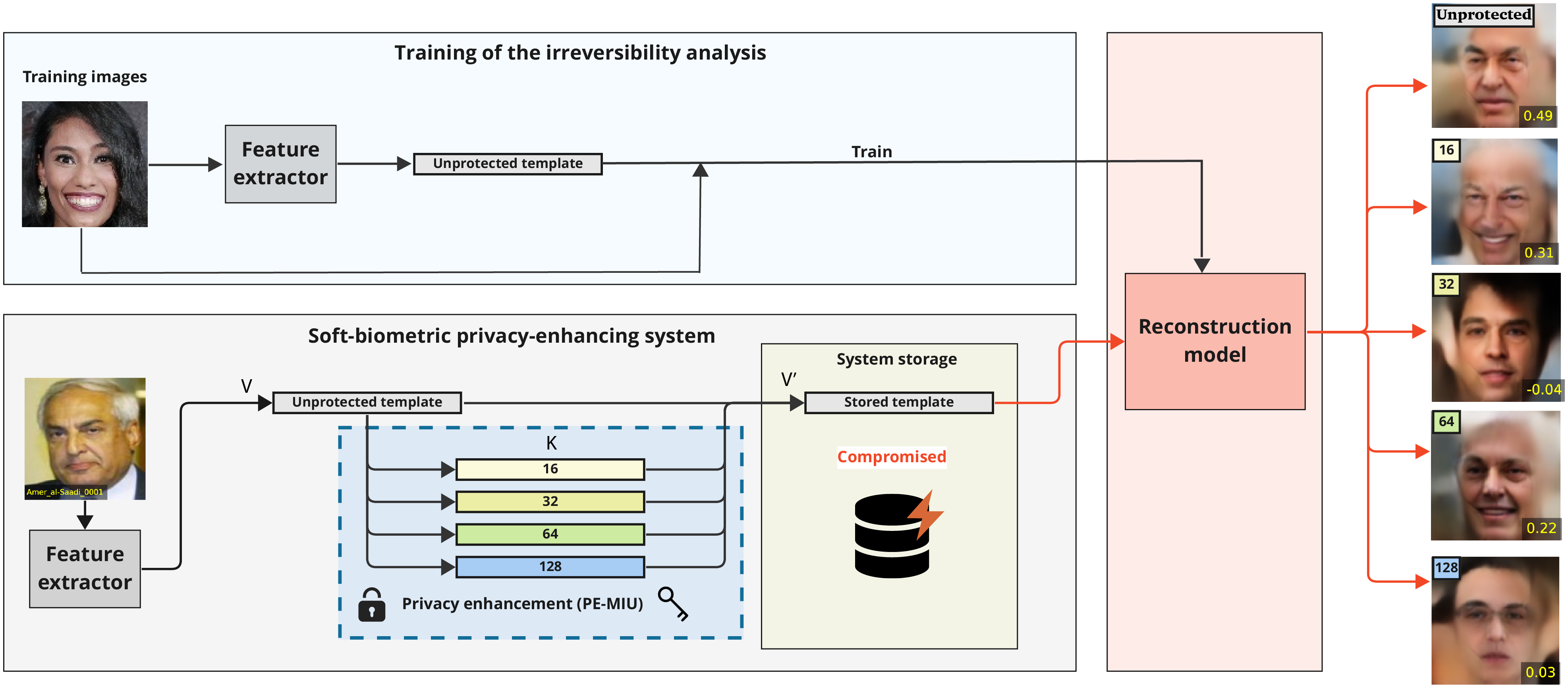}
    \caption{Conceptual overview of the irreversibility analysis of this work. Note that unprotected templates used in the upper part (\ie training process) are different to the unprotected templates that are protected by PE-MIU (\ie bottom part).}
    
    \label{fig:overview}
\end{figure*}

\section{Related works}% and Background}
\label{sec:related-work}

Different works have been proposed with the aim of preventing the derivation of soft-biometric attributes from face embeddings. These approaches have shown good biometric performance while achieving high privacy protection. They are usually focused either on the feature representation (embeddings, \eg~\cite{Terhorst-SoftbiometricEnhancingPE-MIU-2020}) or on the inference-level (comparison, \eg~\cite{Terhorst-NegativeFaceRecognition-2020}). While the former aims at the soft-biometric minimisation (\eg~\cite{Bortolato-SoftbiometricEnhancing-PRFNet-2020}) or the feature transformation (\eg introduction of noise or simple permutation) without excluding the target attribute. In the latter concept, transformations are applied and the biometric comparator is adapted accordingly. In this paper, it is of interest to explore said approaches that introduce transformations at the feature level. As mentioned in Sect.~\ref{sec:intro}, these transformations do not remove soft-biometric information and could therefore leak privacy information. In this context, few works operating at the feature representation have been recently published. Terh{\"o}rst \etal~\cite{Terhorst-NoiseTransformation-2019} proposed a Cosine–Sensitive Noise (CSN) transformation applied to face embeddings to enhance privacy in terms of gender and age attributes. To this end, the authors introduced a specific type of noise over the face representation which hides the soft–biometric information. Further, Terh{\"o}rst \etal~\cite{Terhorst-SoftbiometricEnhancingPE-MIU-2020} proposed a Privacy-Enhancing face recognition approach based on Minimum Information Units (PE-MIU). This method allows the creation of privacy-enhanced face embeddings by partitioning the original embedding into smaller parts (called minimum information units). Then, these blocks are randomly shuffled to obtain a privacy-enhanced embedding. %More recently, Melzi \etal~\cite{Melzi-MultiIVE-WACVW-2023} proposed domain transformation using Principal Component Analysis (PCA), which uses a feature removal strategy using the dimensionality reduction criterion to provide a secure representation across different attributes (gender, age and ethnicity). This approach conducted a comprehensive analysis of the distribution of attributes across the state-of-the-art face embedding extractors. 

The approaches described above have been focused on protecting soft biometric attributes by transforming the feature space. However, these schemes are usually assessed in inadequate evaluations that are not based on standardised protocols~\cite{TerhoerstBIOSIG}. More specifically, it is unknown to what extent such transformed features (\eg~shuffled features) can be reconstructed to their original face images. Moreover, in addition to soft-biometric privacy-enhancement approaches, different privacy-preserving techniques and their integration with theoretical and practical privacy-preservation mechanisms~\cite{Razeghi-CompressedDataSharing-2022,Rezaeifar-PrivacyTemplate-2022,Razeghi-bottlenecks-2023} have been analysed in the literature in terms of recovery and estimation of private attributes. In this paper, we consider a practical privacy-preserving mechanism based on face reconstruction as part of a protocol for evaluating the privacy leakage of soft biometric privacy-enhancing approaches.

\section{Reversing protected deep face templates}
\label{sec:attack-discussion}

Fig.~\ref{fig:overview} shows a conceptual overview depicting the key stages of the analysed attack scenario. For the process of template inversion, an attacker does not have full knowledge of the privacy-protection method used (or the involved random seed), but is in possession of the protected face embeddings (labeled as ``stored template'' on the bottom part in Fig.~\ref{fig:overview}). We based our study on a competitive privacy-protection approach, \ie PE-MIU~\cite{Terhorst-SoftbiometricEnhancingPE-MIU-2020}. In this context of the attack, the non-authorised subject is able to prepare and train an existing DNN-based embedding inversion model to reconstruct face images from unprotected face embeddings (top part in Fig.~\ref{fig:overview}). The same network is subsequently used to reconstruct face images from the protected face embeddings. Note that the protection mechanism of PE-MIU depends on the block size (\ie 16,32,64,128) employed. Given an unprotected face embedding, a random permutation is computed from a previously defined block size. In our work, an irreversibility analysis is performed by comparing the face images reconstructed from the PE-MIU-protected face embeddings with the respective original faces. Finally, the similarity scores computed from the embeddings extracted from the reconstructed face images and their originals are analysed.

\subsection{Soft-biometric privacy protection}
\label{sec:protection}

PE-MIU~\cite{Terhorst-SoftbiometricEnhancingPE-MIU-2020} has been selected as the soft-biometric privacy-enhancement approach since it offers underlying properties that can be easily executed in a realistic attack scenario or adversarial model. As mentioned in Sect.~\ref{sec:related-work}, this method performs a transformation at the feature level preserving entirely the subject's identity without excluding the soft-biometric attribute. Formally, PE-MIU receives a feature embedding of size $S$ containing floating point-based values, denoted by $V \in \mathbb{R}^{S}$. $V$ is then divided into $N=S/K$ blocks of size $K$ representing the minimum of information units that can be transformed into $V$. Further, $N$ blocks of $V$ are randomly exchanged or shuffled, resulting in a new feature vector $V' \in \mathbb{R}^{S}$. Note that the block size $K$ set to transform $V$ functions as a key that is previously defined at application or system level. In other words, the method preserves the same size of blocks and is able to create different transformed vectors $V'$ (\ie different random permutations). Therefore, given $K$, the maximum number of permutations in an authentication or enrolment step would be limited to $N!$.

\subsection{Irreversibility analysis}
\label{sec:irreversibility}

The irreversibility analysis is depicted in the upper part of Fig.~\ref{fig:overview}. In a practical context, it is important to point out that the attacker utilises the previously trained inversion model on features transformed by PE-MIU to reconstruct face images. Subsequently, the attacker can derive the protected soft-biometric attributes from reconstructed embeddings. It is worth mentioning that the main motivation of this work is to estimate the privacy of PE-MIU empirically taking a closer look at one of the privacy requirements for biometric template protection schemes~\cite{ISO-IEC-24745-TemplateProtection-2022}, \ie \textit{irreversibility}. As mentioned in~\cite{Melzi-overview-pets-2022}, the reconstruction of biometric samples similar to the original captured samples from stored protected embeddings is a challenge. However, in the case of PE-MIU, irreversibility may be improved by the permutation complexity or number of shuffled blocks. Here, such permutations can lead to good approximations of the original samples which may reveal soft-biometric information. In an attack scenario, it should be feasible that non-mated comparisons for the same attribute (\eg gender) result in higher similarity scores according to the broad homogeneity effect~\cite{Howard-HomogeneityEffect-2019}, and thus to frequent false matches.

\section{Experimental Setup}
\label{sec:experimenta_setup}

In this section, the implementation details as well as the databases used are outlined in Sect.~\ref{sec:implementation_details}. The main configurations used in the irreversibility analysis are described in Sect.~\ref{sec:irreversibility}. 

\subsection{Implementation details, databases and metrics}
\label{sec:implementation_details}

Two well-known face recognition models are utilised in this work, ArcFace~\cite{Deng-arface-2022} and ElasticFace~\cite{Fadi-elasticface-2022} providing face embeddings of 512-floating point values. Protected embeddings are computed by the approach presented in Sect.~\ref{sec:protection}  using available open-source code\footnote{https://github.com/pterhoer/PrivacyPreservingFaceRecognition}. For the embedding inversion, the approach proposed by~\cite{Shahreza-loss-reconstruction-2022} was selected. Note that this model was previously trained on unprotected face embeddings extracted from the FFHQ database~\cite{Karras-FFHQ-2019} as explained in~\cite{Shahreza-loss-reconstruction-2022}. Subsequently, this trained model is utilised to reconstruct face images from protected face embeddings extracted from the LFW database. We consider a single sample per identity in LFW that obtains the highest quality score computed by the SER-FIQ quality estimator~\cite{Terhorst-SER-FIQ-2020}. Also, the dataset was balanced \wrt the gender attribute resulting in 2,942 unprotected and protected embeddings (identities for each gender attribute), respectively, as it was done in~\cite{OsorioRoig-FaceSoftBiometricPrivacyAttack-TBIOM-2022}. Mated and non-mated comparisons for protected and unprotected embeddings are computed following the protocol View-2 of the LFW database\footnote{https://gitlab.idiap.ch/bob/bob.db.lfw}~\cite{Huang-wild-2007}. Cosine similarity was used as a biometric comparator. Biometric performance is evaluated according to the metrics defined in the international standard ISO/IEC19795-1:2021~\cite{ISO-IEC-19795-1-Framework-210216}: the detection error trade-off curve (DET) comparing the false non-match rate (FNMR) with the false match rate (FMR) as well as equal error rate (EER).

% Biometric performance for protected and unprotected templates was evaluated following the protocol View 2\footnote{https://gitlab.idiap.ch/bob/bob.db.lfw} of the LFW database~\cite{Huang-wild-2007} compliant with the metrics defined in the international standard ISO/IEC19795-1:2021~\cite{ISO-IEC-19795-1-Framework-210216}: the detection error trade-off curve (DET) and equal error rate (EER). 

\begin{figure}[!t]
    \centering
    \subfigure[ArcFace]{\includegraphics[width=0.75\linewidth]{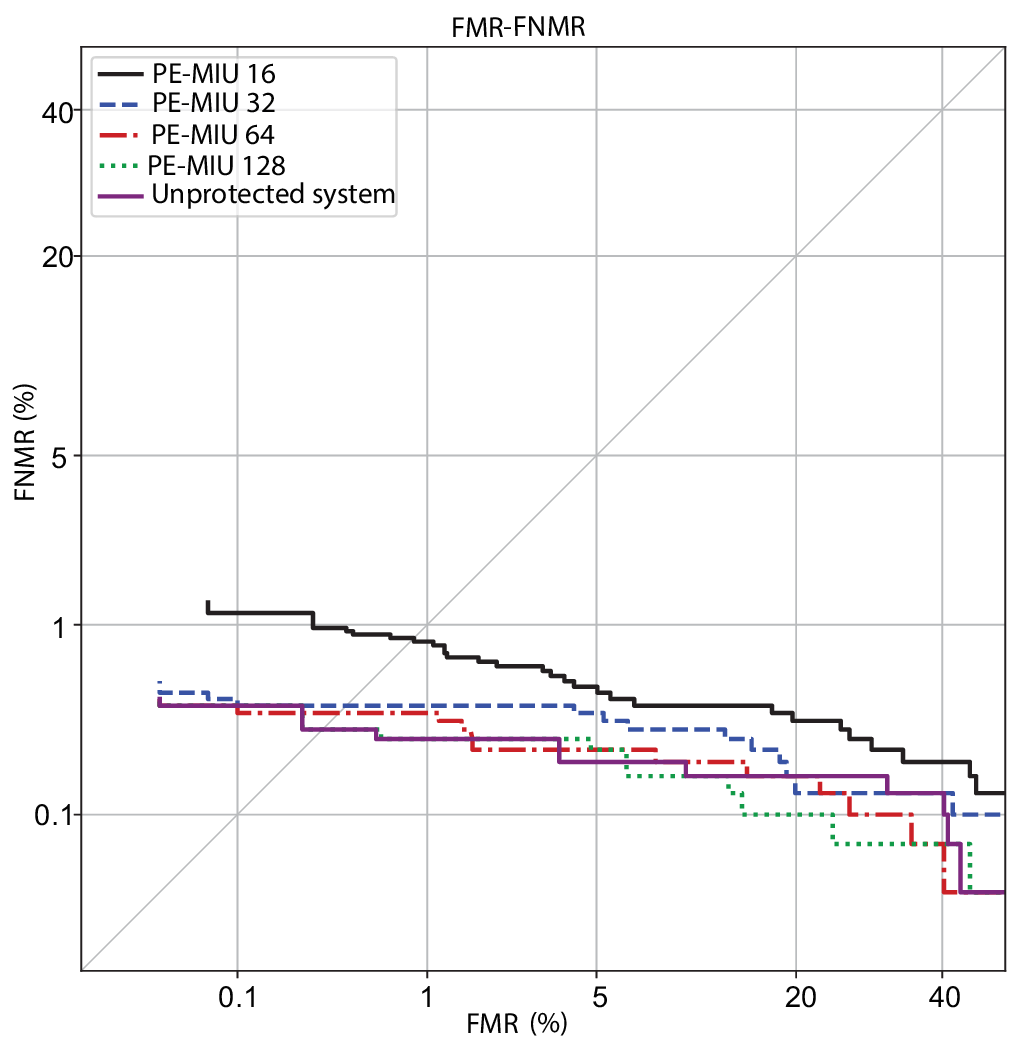}}  
    \hspace{-0.28cm}
    \subfigure[ElasticFace]{\includegraphics[width=0.75\linewidth]{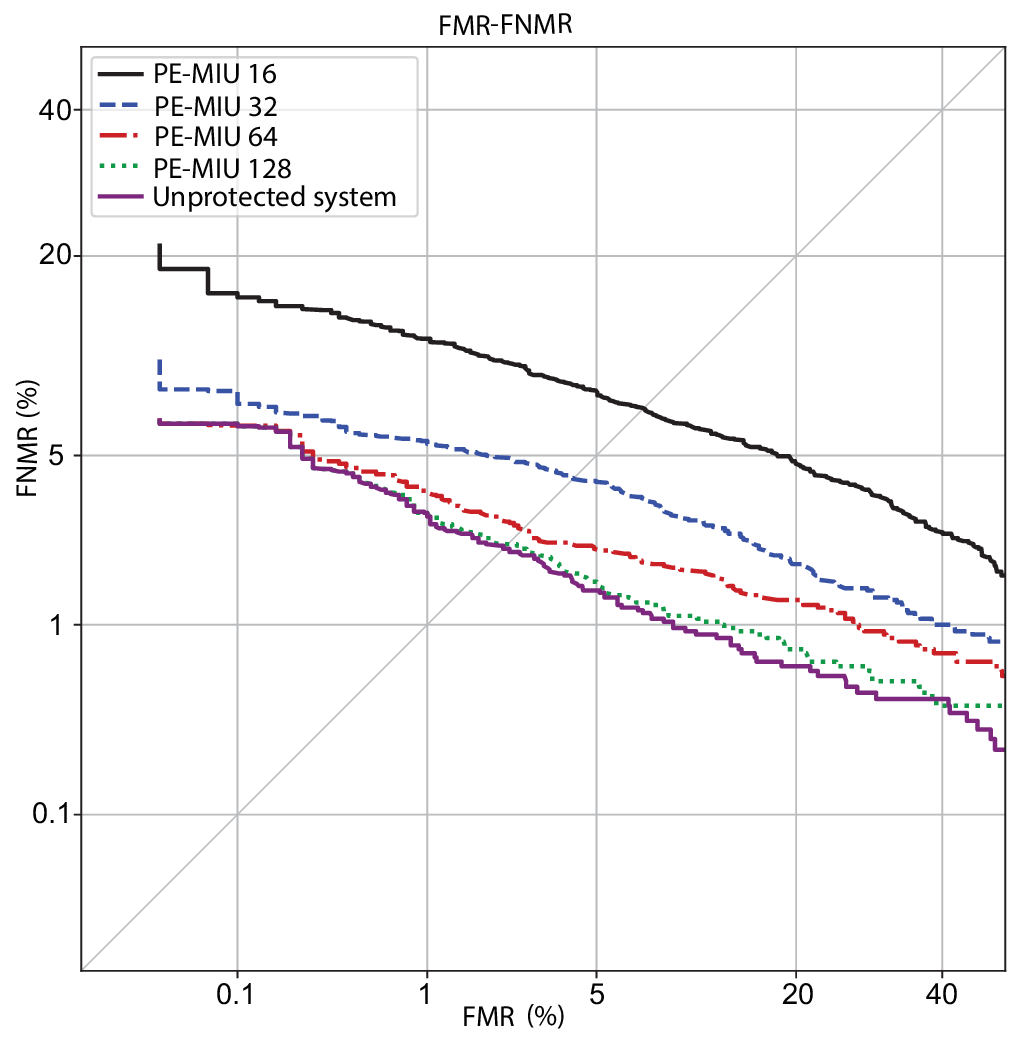}}
    \hspace{-0.28cm}
    \caption{Evaluation of the biometric performance.}
    \label{fig:biometric-performance}
\end{figure}

\subsection{Irreversibility setup}
\label{sec:irreversibility}

For the irreversibility analysis, different block-sizes $K$  are analysed, \ie $K = \{128, 64, 32, 16\}$. In the first set of experiments, we explore the biometric performance, as well as the reversibility success rate (RSR) for different $K$ values (Sect.~\ref{sec:unknow_seeds}). RSR is defined as the ratio of reconstructed face images from protected embeddings that are accepted by the system at a pre-defined decision threshold (\eg at FMR=0.1\%). In the experiments, protected embeddings for a given block-size $K$ are generated (Sect.~\ref{sec:unknow_seeds}). Here, face image reconstruction from different permutations is performed. Furthermore, gender prediction accuracy from reconstructed face embeddings is reported. To that end, traditional Support Vector Machines (SVMs) classifiers are trained on unprotected face embeddings corresponding to the original face images in LFW using different kernels (Poly, RBF, and Sigmoid). Note that the hyperparameters of both classifiers were set to basic configurations without any optimisation and that are used in the training. 

%----before
% For the irreversibility analysis, different block-sizes $K$  are analysed, \ie $K = \{128, 64, 32, 16\}$. In the first set of experiments, we explore the biometric performance, as well as the reversibility success rate (RSR) for different $K$ values (Sect.~\ref{sec:unknow_seeds}). RSR is defined as the ratio of reconstructed face images from protected embeddings that are accepted by the system at a pre-defined decision threshold. In the experiments, protected embeddings for a given block-size $K$ are generated using different unknown random seeds (Sect.~\ref{sec:unknow_seeds}), \ie the attacker has no knowledge about the seed and therefore, different permutations are computed. Furthermore, gender prediction accuracy from reconstructed face embeddings is reported. To that end, traditional Support Vector Machines (SVMs) classifiers are trained on unprotected face embeddings corresponding to the original face images in LFW using different kernels (Poly, RBF, and Sigmoid). Note that the hyperparameters of both classifiers were set to basic configurations without any optimisation and that are used in the training.    

In the second set of experiments, we explore the RSR values for different permutation complexities that can be computed by PE-MIU (Sect.~\ref{sec:know_seeds}). The possible number of permutations ($N!$) for each $K$, \ie $4!$, $8!$, $16!$, and $32!$, is therefore analysed for the considered face embedding extractors. To that end, for a fixed $K$, PE-MIU is utilised so that the same permutation is generated (shuffling the same number of blocks) for all identities. To analyse the full or partial recovery of original face images, protected deep face templates are reconstructed for each $K$ by varying the number of all possible blocks to be shuffled (in this case, $P \in \{2, \ldots, N$\}). Then, the RSR value for each $K$ and $P$ combination is reported.  

% 
% In the second set of experiments, we explore the RSR values for fixed seeds, \ie the attacker has full knowledge of the seeds used by PE-MIU (Sect.~\ref{sec:know_seeds}). The possible number of permutations ($N!$) for each $K$, \ie $4!$, $8!$, $16!$, and $32!$, is therefore analysed. For a fixed $K$ and a given random seed, the same permutation is generated (shuffling the same number of blocks) for all identities. In a practical example, for $K = 128$ with $N! = 4!$ and a random seed that generates a permutation by shuffling only 2 of 4 blocks, all identities would be limited to shuffling only two blocks of their embeddings. To analyse the full or partial recovery of original face images, protected deep face templates are reconstructed for each $K$ by varying the number of all possible blocks to be shuffled (in this case, $P \in \{2, \ldots, N$\}). Then, the RSR value for each $K$ and $P$ combination is reported.  

% It is worth noting that reconstructed face images are always compared with their corresponding original faces in the above experiments. Therefore, this type of evaluation analyses whether the image reconstructed from a protected template will be accepted against the same sample representing the unprotected template according to a threshold defined in the system. 

%table baseline of face reconstruction when random seed is not fixed
\begin{table}[!t]
	\caption{Error rates(\%) reported on thresholds fixed at the biometric system.}
	\label{tab:error-rates}
    \begin{adjustbox}{max width=\linewidth}
    \begin{tabular}{ccc cccc}
    \toprule
        \textbf{Model}&\textbf{Protection} & \textbf{EER}& \multicolumn{2}{c}{\textbf{FMR=0.1}}&\multicolumn{2}{c}{\textbf{FMR=1.0}}  \\ \cmidrule{4-7} 
                       &                     &             &\textbf{FNMR}&\textbf{RSR}    &  \textbf{FNMR}&\textbf{RSR} \\ \cmidrule{1-7}

    \multirow{5}{*}{ArcFace}& Unprotected & 0.30&0.40&\textbf{100.00}&0.27&\textbf{100.00} \\

                            &16           &0.87&1.13&0.11&0.83&2.12 \\
                            &32           &0.40&0.40&0.60&0.40&4.66 \\
                            &64           &0.37&0.37&4.62&0.37&14.48 \\
                            &128          &0.30&0.40&19.22&0.27&34.65 \\ \cmidrule{1-7}

    \multirow{5}{*}{ElasticFace}& Unprotected &2.17&6.33&\textbf{99.63}&2.93&\textbf{99.82} \\
                                & 16          &7.27&15.67&0.08&12.00&1.04 \\
                                & 32          &4.10&7.53&0.59&5.50&3.80 \\
                                & 64          &2.63&6.37&5.32&3.63&15.42 \\
                                &128          &2.27&6.33&23.94&3.00&39.48 \\

     \bottomrule
     \end{tabular}
     \end{adjustbox}

\end{table}

%table baseline of gender prediction when random seed is not fixed
\begin{table}[!t]
	\caption{Gender prediction accuracy.}
	\label{tab:gender-prediction}
    \begin{adjustbox}{max width=\linewidth}
    \begin{tabular}{cc ccc}
    \toprule
        \multirow{2}{*}{\textbf{Model}}&\multirow{2}{*}{\textbf{Protection}} & \multicolumn{3}{c}{\textbf{SVM}}  \\ \cmidrule{3-5}
                       &                     &             \textbf{Poly}&\textbf{RBF}&\textbf{Sigmoid} \\ \cmidrule{1-5}

                      \multirow{5}{*}{ArcFace}&Unprotected&\textbf{0.81} $\pm$ 0.02 & \textbf{0.89} $\pm$ 0.01 & \textbf{0.85} $\pm$ 0.02  \\
                                              &16          &0.50 $\pm$ 0.03 & 0.50 $\pm$ 0.03 & 0.49 $\pm$ 0.02 \\
                                              &32          &0.50 $\pm$ 0.03 & 0.50 $\pm$ 0.03 & 0.49 $\pm$ 0.02  \\
                                              &64          &0.52 $\pm$ 0.02 & 0.53 $\pm$ 0.03 & 0.52 $\pm$ 0.03 \\
                                              &128         & 0.55 $\pm$ 0.02 & 0.57 $\pm$ 0.02 & 0.56 $\pm$ 0.01 \\ \cmidrule{1-5}
                     \multirow{5}{*}{ElasticFace}&Unprotected&\multicolumn{1}{l}{\textbf{0.85} $\pm$ 0.02} & \multicolumn{1}{l}{\textbf{0.88} $\pm$ 0.02} & \multicolumn{1}{l}{\textbf{0.84} $\pm$ \textbf{0.02}} \\
                                                &16& \multicolumn{1}{l}{0.52 $\pm$ 0.02} & \multicolumn{1}{l}{0.52 $\pm$ 0.03} & \multicolumn{1}{l}{0.51 $\pm$ 0.03} \\
                                                
                                                &32&  \multicolumn{1}{l}{0.52 $\pm$ 0.02} & \multicolumn{1}{l}{0.55 $\pm$ 0.02} & \multicolumn{1}{l}{0.54 $\pm$ 0.03} \\
                                                
                                                &64& \multicolumn{1}{l}{0.57 $\pm$ 0.01} & \multicolumn{1}{l}{0.58 $\pm$ 0.03} & \multicolumn{1}{l}{0.57 $\pm$ 0.03} \\
                                                
                                                &128& \multicolumn{1}{l}{0.64 $\pm$ 0.03} & \multicolumn{1}{l}{0.64 $\pm$ 0.02} & \multicolumn{1}{l}{0.63 $\pm$ 0.03} \\

     \bottomrule
     \end{tabular}
     \end{adjustbox}

\end{table}

\section{Experimental Results}
\label{sec:results}

\subsection{Face image reconstruction from protected templates}
\label{sec:unknow_seeds}

First, the biometric performance between the unprotected face embeddings and their corresponding protected embeddings for different values of $K$ is shown in Fig.~\ref{fig:biometric-performance}. In this experiment, different random permutations are generated for each identity by PE-MIU. Note that the biometric performance improves with the block-size ($K$) for both face recognition systems. However, higher values of $K$ are assumed to lead to lower privacy protection~\cite{Terhorst-SoftbiometricEnhancingPE-MIU-2020}. 

%--> before
% First, the biometric performance between the unprotected face embeddings and their corresponding protected embeddings for different values of $K$ is shown in Fig.~\ref{fig:biometric-performance}. In this experiment, a random seed has not been set in PE-MIU. Therefore, different random permutations are generated for each identity. Note that the biometric performance improves with the block-size ($K$) for both face recognition systems. However, higher values of $K$ are assumed to lead to lower privacy protection~\cite{Terhorst-SoftbiometricEnhancingPE-MIU-2020}. 

Tab.~\ref{tab:error-rates} reports the RSRs of face images reconstructed from unprotected and protected embeddings, respectively. As expected, the highest reversibility chances are achieved for unprotected embeddings (RSRs $\geq 99.63\%$), while for the protected embeddings, they are in the ranges of 0.08\% to 39.48\% depending on $K$. %Note that the non-authorised subject requires zero effort lower bounded by the probability of guessing the random permutation. The latter scenario would be in a context of attack where the attacker requires zero effort and obviously, the random seed is unknown. 
Tab.~\ref{tab:gender-prediction} confirms the results presented in Tab.~\ref{tab:error-rates}: a low probability of gender prediction among the different classifiers as a function of $K$ is perceived.

%table rversing according to permutation complexity for FMR=0.1
\begin{table}[!t]
	\caption{Reversibility success rates - RSRs (in \%) for different combinations $K$ and $P$ on the two face recognition models at different security thresholds.}
	\label{tab:reversing-arcface}
    \begin{adjustbox}{max width=\linewidth}
    \begin{tabular}{cc cc cc}
    \toprule
        \multirow{2}{*}{\textbf{Block size ($\mathbf{K}$)}}&\multirow{2}{*}{\textbf{P}}  & \multicolumn{2}{c}{\textbf{ArcFace}}     & \multicolumn{2}{c}{\textbf{ElasticFace}} \\ 
                   &            &  \textbf{FMR=0.1\%} & \textbf{FMR=1.0\%} & \textbf{FMR=0.1\%} & \textbf{FMR=1.0\%}   \\          
                               \midrule
        
        % \multirow{14}{*}{32}   &                                                                      0& 95.04&99.56\\
         \multirow{13}{*}{32}                      &                                                  4& 89.09      & 96.64     &97.83   &   99.86\\
                               &                                                                      5& 84.06      & 95.45     & 96.67  &   99.66\\
                               &                                                                      6& 78.59      & 93.51     &92.39   &   99.15\\
                               &                                                                      7& 70.85      & 90.59     &87.09   &   98.37\\
                               &                                                                      8& 58.44      & 85.39     &75.77   &   95.65\\
                               &                                                                      9& 43.56      & 76.18     & 59.97  &  89.26\\
                               &                                                                      10& 32.79     & 67.01     & 44.10  &  79.95\\
                               &                                                                      11& 20.42     & 52.97     & 25.04  & 63.81\\
                               &                                                                      12& 9.68      & 36.29     & 10.26  & 43.39\\
                               &                                                                      13& 3.67      & 20.93     & 3.26   & 22.26\\
                               &                                                                      14& 1.39      & 9.85      & 0.85   & 7.37\\
                               &                                                                      15& 0.44      & 4.35      & 0.17   & 2.48\\
                               &                                                                      16& 0.0       & 1.39      & 0.00   & 0.48\\ \midrule
    
       % \multirow{9}{*}{64}     &                                                                      0& 94.6& 99.73\\
         \multirow{7}{*}{64}   &                                                                      2& 88.96      &  96.33    & 98.03  & 99.69\\
                               &                                                                      3& 78.49      &  93.24    & 92.05  & 99.15\\
                               &                                                                      4& 60.45      &  87.53    & 75.33  & 95.17\\
                               &                                                                      5& 30.82      &  65.14    & 40.27  & 78.63\\
                               &                                                                      6& 9.58       &  35.37    & 9.0    & 38.63\\
                               &                                                                      7& 1.22       &  10.50    & 0.78   & 9.04\\
                               &                                                                      8& 0.07       &  1.16     & 0.00   & 0.37\\ \midrule

% \multirow{4}{*}{128}           &                                                                        0& 94.6& 99.73\\
\multirow{3}{*}{128}           &                                                                     2& 60.58       &  85.93    & 75.94  & 95.62\\
                               &                                                                     3& 8.83        &  32.76    & 8.97   & 40.13\\
                               &                                                                     4& 0.03        &  1.19     & 0.00   & 0.37\\

     \bottomrule
     \end{tabular}
     \end{adjustbox}
\end{table}

\begin{figure*}[!t]
    \centering
     \subfigure{\includegraphics[width=\linewidth]{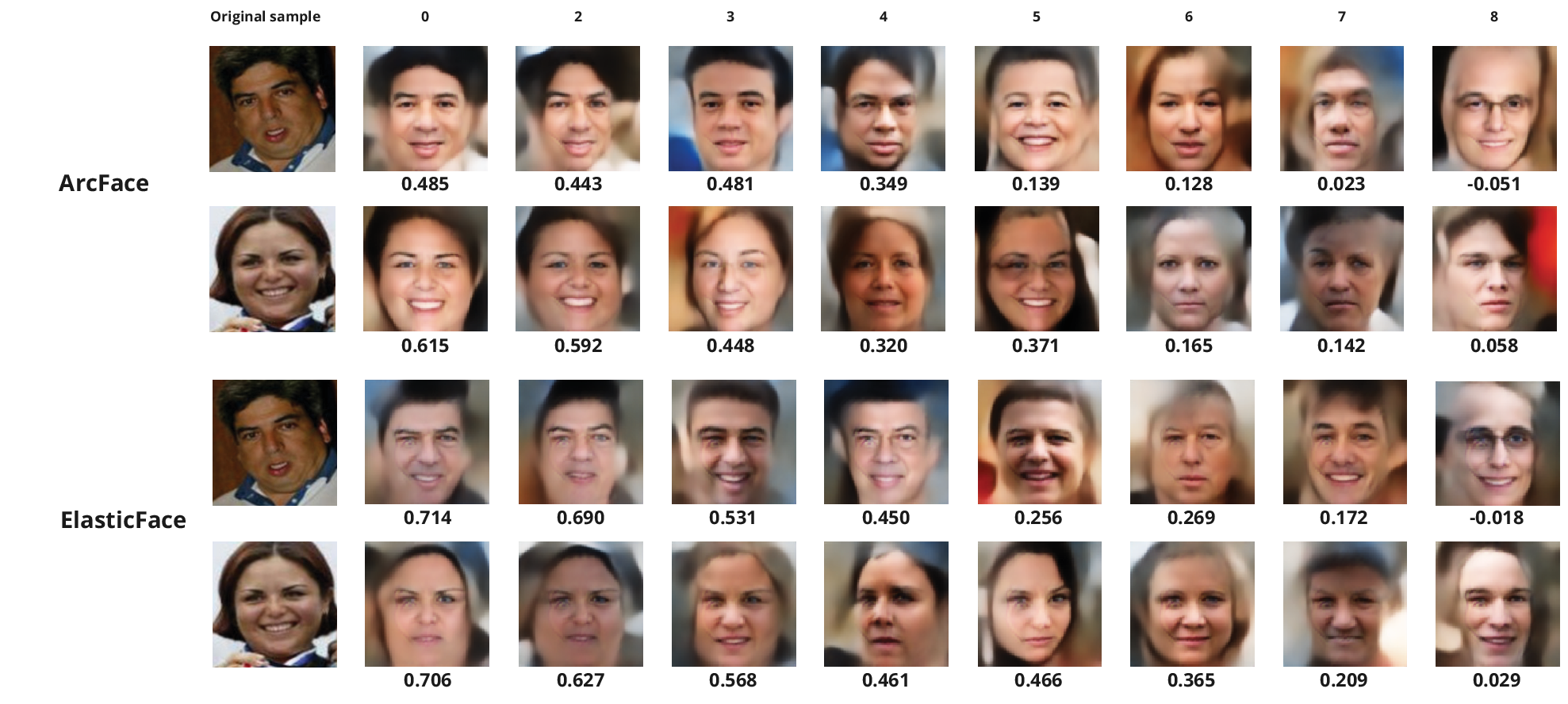}}  
    \caption{Examples of face images reconstructed from protected deep face embeddings for different values $P$ (\ie $2\leq N \leq 8$) using PE-MIU with block size of $K=64$. Note that the value zero represents the face reconstruction from the unprotected embedding, where no blocks in the vector were swapped. Similarity scores are shown by comparing face images reconstructed from protected face embeddings with their corresponding face embeddings extracted from original face images, shown in the left column.}
    \label{fig:faces-reconstruction-examples}
\end{figure*}

\subsection{Detailed analysis on the permutation complexity}
\label{sec:know_seeds}

We report in Tab.~\ref{tab:reversing-arcface} the RSR value for reversing face images from face embeddings in a scenario where the same permutation complexity (\ie $P$) is retained for all protected templates. Note that some configurations are omitted where no change in the observed trend is expected, \eg $K = 16$. We observe that the RSR decreases as $P$ increases for all values of $K$  across different security thresholds. In particular, the probability of successfully inverting a protected face embedding for different values of $K$ is above 85.39\% and 95.17\% for ArcFace and ElasticFace, respectively, at FMR = 1.0\%, when at least half of the blocks are shuffled. A reduction in RSR down to 58.44\% and 75.33\% is perceived for both recognition systems for a high safety threshold, \ie FMR=0.1\%. Note the vulnerability of the ElasticFace face recognition model in comparison to ArcFace: the highest RSRs yielded by ElasticFace are about 99\% and 97\% for FMR=1.0\% and FMR=0.1\%, respectively, while ArcFace achieves approximately 96\% and 89\% for the same thresholds. It should be also observed that the results attained for lower $P$ values are directly proportional to the $K$ values, \ie RSR improves as $P$ and $K$ drop. This observation could be explained due to the minimum information (entropy) that is managed across the blocks in the shuffling process which offers ``useful'' guesses of the floating-point values in the face reconstruction process. Fig~\ref{fig:faces-reconstruction-examples} shows examples of faces reconstructed from different permutations that also reveal gender information. It should be noted that the analysis in terms of block size and the number of shuffled blocks provided by the unique design of PE-MIU motivates further works (\eg information bottleneck models~\cite{Razeghi-CompressedDataSharing-2022}) to explore which features should be retained or suppressed to avoid privacy leakage in the considered embedding space, in particular, in face embeddings of size 512.

\subsection{Discussion on random seeds}
\label{sec:successful-attack}

It is important to note that the transformations explained in Sect.~\ref{sec:protection} and evaluated in Sect.~\ref{sec:unknow_seeds} are usually generated from a pseudo-random number generator stored in \eg a physical device. Thus, for a specific application, a transformation (\ie a permutation) applied to a feature vector would be computed through a seed produced by a pseudo-random number generator. In this work, a scenario without seed specification has been evaluated (Sect.~\ref{sec:unknow_seeds}). However, it is worth noting that an adversary or attacker with full knowledge of the system's random seed could easily manipulate PE-MIU. In particular, the attacker could reverse the permutation process by re-shuffling blocks based on the known seed. Finally, it would be possible to reconstruct face images from a fully inverted protected template. In this case, RSR values are expected to be approximately 100\%, similar to the reconstruction of face images from unprotected embeddings presented in Sect~\ref{sec:unknow_seeds}. In addition, an attacker could launch a brute-force attack on the random seed used by the PE-MIU method. Here, the attacker's effort has to be taken into account, since a high permutation complexity (\ie a large number of shuffled blocks) would have to be guessed.

\section{Summary}
\label{sec:remarks}

In this work, it was empirically demonstrated that face embeddings protected by soft-biometric privacy-enhancement schemes may be successfully reconstructed, depending on certain parameters of the privacy-enhancement approach (\eg $K$ defined by PE-MIU and a permutation complexity $P$). Experimental evaluations conducted on an open-source privacy-enhancement method and a freely available database, \ie LFW, showed that the reversibility success rates (RSRs) are low in a scenario where the attacker does not have any knowledge about employed random seeds. A detailed analysis of the transformation complexity used by the protection mechanism showed that RSR values can increase up to 95.62\%. In experiments, face image were reconstructed that achieved high similarity scores with respect to their corresponding original face images for different state-of-the-art face recognition models. Generally, high RSRs were obtained when transformations exhibit low complexity or an attacker knows or is able to guess a random seed. Future work will be focused on a deeper analysis that considers the reversibility and reconstruction of protected face embeddings as part of a joint training process across different soft-biometric privacy-enhancement approaches.

% In this work, it was empirically demonstrated that deep face templates with soft biometric attribute protection can be successfully reconstructed, depending on certain parameters of the privacy-enhancement approach (\ie $K$, $P$, and knowledge of a random seed defined by PE-MIU). Experimental evaluation conducted on an open-source privacy-enhancement method and a freely available database, \ie LFW, showed that the reversibility success rates (RSRs) increased for a high block-size, resulting in an RSR of 39.48\% for $K = 128$ at FMR = 1.0\% when a random seed is unknown, while, on the other hand, RSRs were increased up to 95.62\% in a scenario where the random seed is known. The experiments led to face image reconstructions that achieved high similarity scores with respect to their corresponding original face images for different state-of-the-art face recognition models. More precisely, high RSRs were perceived when transformations for protecting the attribute at the feature level provided a low $P$ value in a scenario where the random seed is known. 

%\section*{Acknowledgements}
%\label{sec:acknowledgements}
%This work has in part received funding from the European Union’s Horizon 2020  research and innovation programme under the Marie Skłodowska-Curie grant agreement No. 860813 - TReSPAsS-ETN and the German Federal Ministry of Education and Research and the Hessen State Ministry for Higher Education, Research and the Arts within their joint support of the National Research Center for Applied Cybersecurity ATHENE.
{\small
\bibliographystyle{IEEEtran}
\bibliography{IEEEabrv}
}

\end{document}